\def\year{2022}\relax
\documentclass[letterpaper]{article} 
\usepackage{aaai22}  
\usepackage{times}  
\usepackage{helvet}  
\usepackage{courier}  
\usepackage[hyphens]{url}  
\usepackage{graphicx} 
\usepackage{natbib}  
\usepackage{caption} 
\DeclareCaptionStyle{ruled}{labelfont=normalfont,labelsep=colon,strut=off} 
\usepackage{algorithm}
\usepackage[noend]{algorithmic}

\usepackage{newfloat}
\usepackage{listings}
\floatstyle{ruled}
\newfloat{listing}{tb}{lst}{}
\floatname{listing}{Listing}

\usepackage{amssymb}
\usepackage{amsmath}
\usepackage{multirow}
\usepackage{booktabs}
\usepackage{array}
\usepackage{arydshln}
\usepackage{pgfplots, pgfplotstable}

\pgfplotsset{
    compat=1.17
} 
\usepgfplotslibrary{groupplots}

\usepackage{tikz}

\usepackage{CJKutf8}
\usepackage{color}

\title{Unified Named Entity Recognition as Word-Word Relation Classification}

\author {
    Jingye Li,\textsuperscript{\rm 1,}$^*$
    Hao Fei,\textsuperscript{\rm 1,}\footnote{Equal contribution}
    Jiang Liu,\textsuperscript{\rm 1}
    Shengqiong Wu,\textsuperscript{\rm 1} \\
    Meishan Zhang,\textsuperscript{\rm 2}
    Chong Teng,\textsuperscript{\rm 1}
    Donghong Ji,\textsuperscript{\rm 1}
    Fei Li\textsuperscript{\rm 1,}\thanks{Corresponding author}
}
\affiliations {
    \textsuperscript{\rm 1} Key Laboratory of Aerospace Information Security and Trusted Computing, Ministry  \\of Education,
School of Cyber Science and Engineering, Wuhan University, China \\
    \textsuperscript{\rm 2} Institute of Computing and Intelligence, Harbin Institute of Technology (Shenzhen), China\\
    \{theodorelee, hao.fei, liujiang, whuwsq, tengchong, dhji, lifei\_csnlp\}@whu.edu.cn, mason.zms@gmail.com
}

\usepackage{bibentry}

\begin{document}
\begin{CJK}{UTF8}{gbsn}
\maketitle

\begin{abstract}
So far, named entity recognition (NER) has been involved with three major types, including flat, overlapped (aka. nested), and discontinuous NER, which have mostly been studied individually. 
Recently, a growing interest has been built for unified NER, tackling the above three jobs concurrently with one single model.
Current best-performing methods mainly include span-based and sequence-to-sequence models, where unfortunately the former merely focus on boundary identification and the latter may suffer from exposure bias.
In this work, we present a novel alternative by modeling the unified NER as word-word relation classification, namely W$^2$NER.
The architecture resolves the kernel bottleneck of unified NER by effectively modeling the neighboring relations between entity words with \texttt{Next-Neighboring-Word} (NNW) and \texttt{Tail-Head-Word-*} (THW-\texttt{*}) relations.
Based on the W$^2$NER scheme we develop a neural framework, in which the unified NER is modeled as a 2D grid of word pairs.
We then propose multi-granularity 2D convolutions for better refining the grid representations.
Finally, a co-predictor is used to sufficiently reason the word-word relations.
We perform extensive experiments on 14 widely-used benchmark datasets for flat, overlapped, and discontinuous NER (8 English and 6 Chinese datasets), where our model beats all the current top-performing baselines, pushing the state-of-the-art performances of unified NER.\footnote{Codes available at \url{https://github.com/ ljynlp/W2NER.git}}
\end{abstract}

\section{Introduction}

\begin{figure}
    \centering
    \includegraphics[width=0.85\columnwidth]{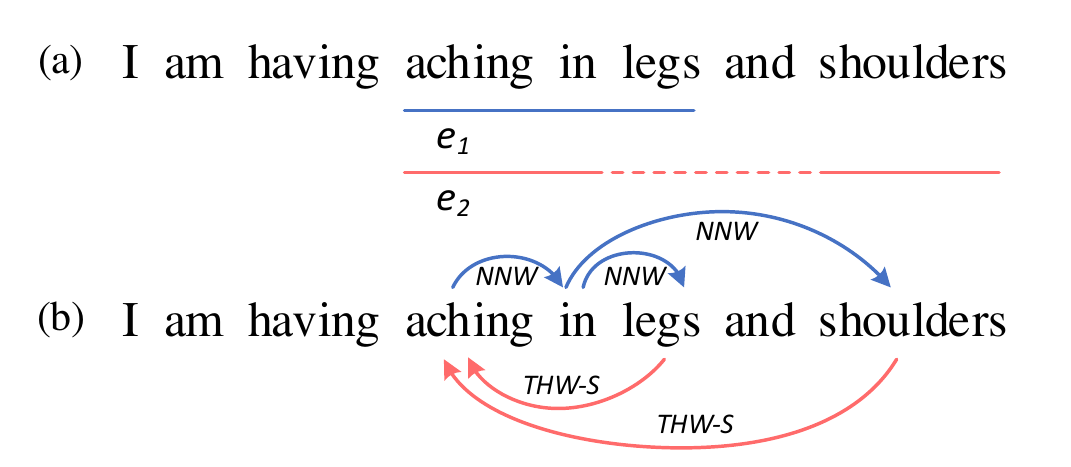}
    \caption{
    (a) An example to show three types of NER. 
    $e_1$ is a flat entity overlapped with a discontinuous entity $e_2$ at the span ``aching in''. 
    (b) We formalize three NER subtasks as word-word relation classification, where the \texttt{Next-Neighboring-Word} (NNW) relation indicates that a word pair are successively joint as a segment of an entity (e.g., aching$\rightarrow$in), 
    and the \texttt{Tail-Head-Word-*} (THW-\texttt{*}) relation implies the edges where the tail words connect to the head words (e.g., legs$\rightarrow$aching) as an entity with ``\texttt{*}'' type (e.g., \emph{Symptom}). 
    }
    \label{fig:example}
\end{figure}

Named entity recognition (NER) has long been a fundamental task in natural language processing (NLP) community, due to its wide variety of knowledge-based applications, e.g., relation extraction \cite{wei2020novel,li2021mrn}, entity linking \cite{le2018improving,hou2020improving}, etc. 
Studies of NER have gradually evolved initially from the flat NER \cite{lample2016neural,strubell2017fast}, late to the overlapped NER \cite{yu2020named,shen2021locate}, and recently to the discontinuous NER \cite{dai2020effective,li2021span}.
Specifically, flat NER simply detects the mention spans and their semantic categories from text, while the problems in overlapped and discontinuous NER become more complicated, i.e., overlapped entities contain the same tokens,\footnote{
Without losing generality, ``nested'' can be seen as a special case of ``overlapped''  \cite{zeng-etal-2018-extracting,dai-2018-recognizing,fei2020boundaries}. } and discontinuous entities entail non-adjacent spans, as illustrated in Figure \ref{fig:example}.

Previous methods for multi-type NER can be roughly grouped into four major categories: 1) \textit{sequence labeling}, 2) \textit{hypergraph-based methods}, 3) \textit{sequence-to-sequence methods} and 4) \textit{span-based methods}.
A majority of initial work formalizes NER as a sequence labeling problem \cite{lample2016neural,zheng-etal-2019-boundary,tang2018recognizing,strakova2019neural}, assigning a tag to each token.
However, it is difficult to design one tagging scheme for all NER subtasks.
Then hypergraph-based models are proposed \cite{lu2015joint,wang2018neural,katiyar2018nested} to represent all entity spans,
which however suffer from both the spurious structure and structural ambiguity issue during inference. 
Recently, \citet{yan2021unified} propose a sequence-to-sequence (Seq2Seq) model to directly generate various entities, which unfortunately potentially suffers from the decoding efficiency problem and certain common shortages of Seq2Seq architecture, e.g., exposure bias.
Span-based methods \cite{luan2019general,li2021span} are another state-of-the-art (SoTA) approaches for unified NER, enumerating all possible spans and conduct span-level classification.
Yet the span-based models can be subject to maximal span lengths and lead to considerable model complexity due to the enumerating nature.
Thus, designing an effective unified NER system still remains challenging.

Most of the existing work has paid the major focus on how to accurately identify the entity boundary, i.e., the kernel problem of NER, especially for flat one \cite{strakova2019neural,fei2021rethinking}.
However, after carefully rethinking the common characteristics of all three types of NER, we find that the bottleneck of unified NER more lies in the modeling of the neighboring relations between entity words.
Such adjacency correlations essentially describe the semantic connectivity between the partial text segments, which especially plays the key role for the overlapping and discontinuous ones.
As exemplified in Figure \ref{fig:example}(a), it could be effortless to detect the flat mention ``aching in legs'', since its constituent words all are naturally adjacent.
But, to detect out the discontinuous entity ``aching in shoulders'', effectively capturing the semantic relations between the neighboring segments of ``aching in'' and ``shoulders'' is indispensable.

On the basis of the above observation, we in this paper investigate an alternative unified NER formalism with a novel word-word relation classification architecture, namely W$^2$NER.
Our method resolves the unified NER by effectively modeling both the entity boundary identification as well as the neighboring relations between entity words.
Specifically, W$^2$NER makes predictions for two types of relations, including the \texttt{Next-Neighboring-Word} (NNW) and the \texttt{Tail-Head-Word-*} (THW-\texttt{*}), as illustrated in Figure \ref{fig:example}(b).
The NNW relation addresses entity word identification, indicating if two argument words are adjacent in an entity (e.g., aching$\rightarrow$in), 
while the THW-\texttt{*} relation accounts for entity boundary and type detection, revealing if two argument words are the tail and head boundaries respectively of ``\texttt{*}'' entity (e.g., legs$\rightarrow$aching, \emph{Symptom}).

\begin{figure*}[t!]
\centering
\includegraphics[width=1.0\textwidth]{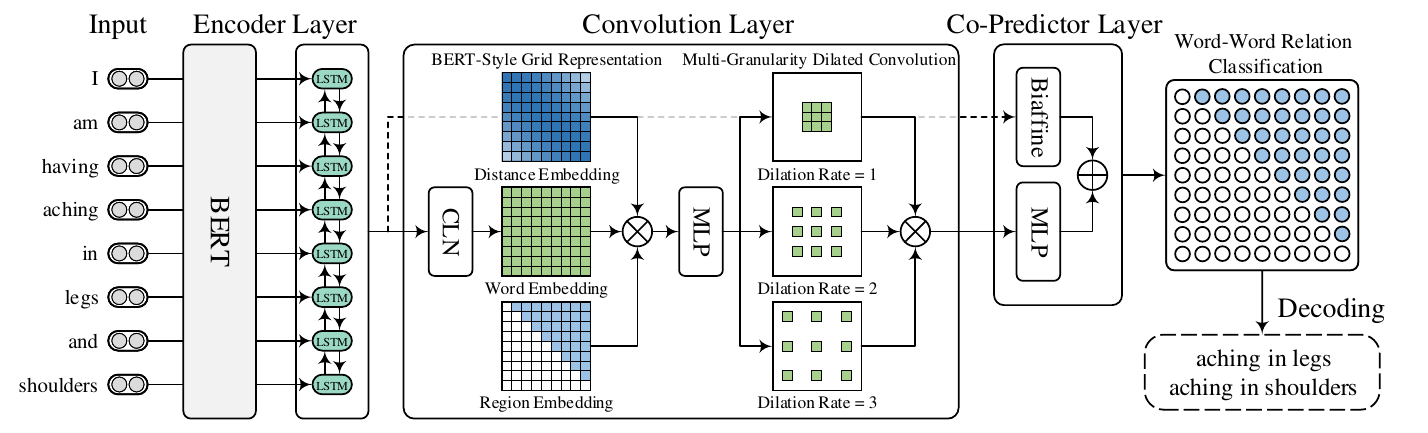}
\caption{
Overall NER architecture.
CLN and MLP represent conditional layer normalization and multi-layer perceptron.
{\small $\bigoplus$} and {\small $\bigotimes$} represent element-wise addition and concatenation operations. 
}
\label{fig:architecture}
\end{figure*}

Based on the W$^2$NER scheme, we further present a neural framework for unified NER (cf. Figure \ref{fig:architecture}).
First, BERT \cite{devlin2018bert} and BiLSTM \cite{lample2016neural} are used to provide contextualized word representations, based on which we construct a 2-dimensional (2D) grid for word pairs.
Afterwards, we design multi-granularity 2D convolutions to refine the word-pair representations, effectively capturing the interactions between both the close and distant word pairs. 
A co-predictor finally reasons the word-word relations and produces all possible entity mentions, in which the biaffine and the multi-layer perceptron (MLP) classifiers are jointly employed for the complementary benefits.

We conduct extensive experiments on 14 datasets, ranging from 2 English and 4 Chinese datasets for flat NER, 3 English and 2 Chinese datasets for overlapped NER, 3 English datasets for discontinuous NER.
Compared with 12 baselines for flat NER, 7 baselines for overlapped NER, 7 baselines for discontinuous NER, our model achieves the best performances on all the datasets, becoming the new SoTA method of unified NER.
Our contributions include:

$\bullet$ We present an innovative method that casts unified NER as word-word relation classification, where both the relations between boundary-words and inside-words of entities are fully considered.

$\bullet$ We develop a neural framework for unified NER, in which we newly propose a multi-granularity 2D convolution method for sufficiently capturing the interactions between close and distant words.

$\bullet$ Our model pushes current SoTA performances of NER on total 14 datasets.

\begin{figure}
    \centering
    \includegraphics[width=0.8\columnwidth]{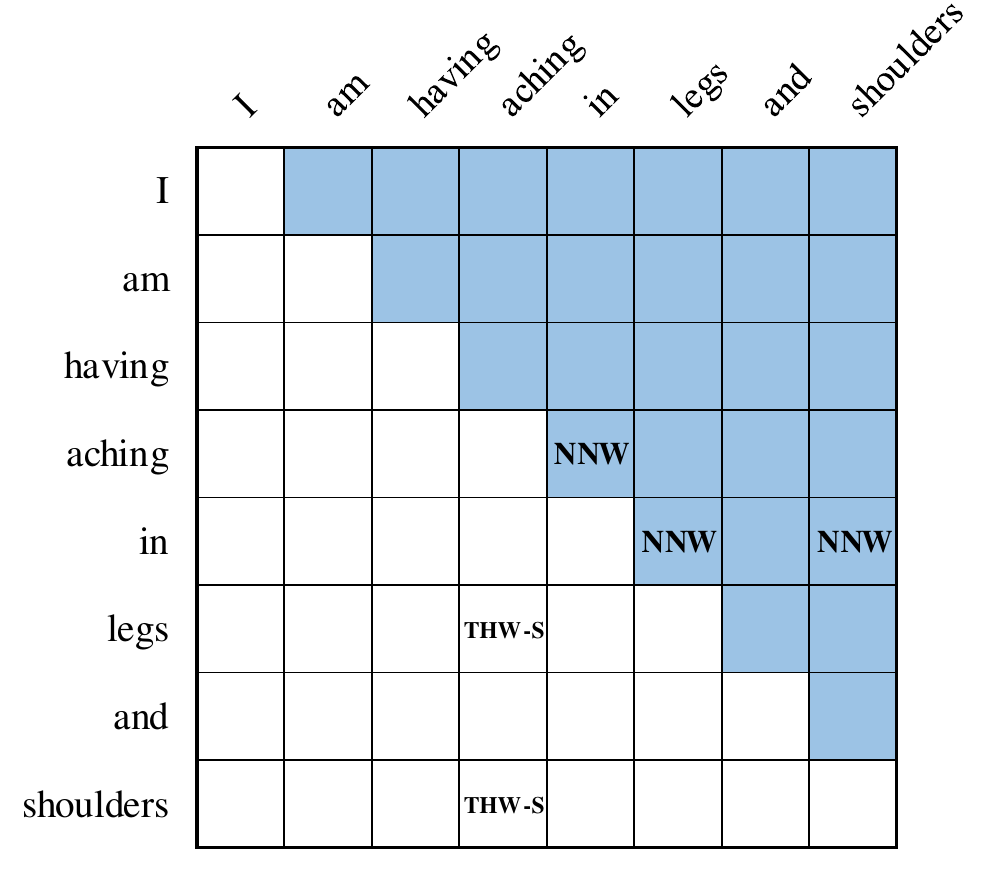}
    \caption{An example to show our relation classification method for NER. We leverage a word-pair grid to visualize the relations between each word pair. NNW denotes the \texttt{Next-Neighboring-Word} relation and THW-S denotes the \texttt{Tail-Head-Word} relation that exists in a ``Symptom'' entity.
    To avoid the sparsity of relation instances, NNW and THW relations are tagged in the upper and lower triangular regions.}
    \label{fig:grid}
\end{figure}

\section{NER as Word-Word Relation Classification}
Flat, overlapped, discontinuous NER can be formalized as follows:
given an input sentence consisting of $N$ tokens or words $X = \{x_1, x_2, ..., x_N\}$,
the task aims to extract the relations $\mathcal{R}$ between each token pairs $(x_i, x_j)$,
where $\mathcal{R}$ is pre-defined, including \texttt{NONE}, \texttt{Next-Neighboring-Word} (NNW), and \texttt{Tail-Head-Word-*} (THW-\texttt{*}).
These relations can be explained as below and we also give an example as demonstrated in Figure \ref{fig:grid} for better understanding.

\begin{itemize}
\setlength{\topsep}{0pt}
\setlength{\itemsep}{0pt}
\setlength{\parsep}{0pt}
\setlength{\parskip}{0pt}
  \item \texttt{NONE}, indicating that the word pair does not have any relation defined in this paper.
  \item \texttt{Next-Neighboring-Word}: the NNW relation indicates that the word pair belongs to an entity mention, and the word in certain row of the grid has a successive word in certain column of the grid. 
  \item \texttt{Tail-Head-Word-*}: the THW relation indicates that the word in certain row of the grid is the tail of an entity mention, and the word in certain column of the grid is the head of an entity mention. ``\texttt{*}'' indicates the entity type.
\end{itemize}

With such design, our framework is able to identify flat, overlapped and discontinuous entities simultaneously.
As shown in Figure \ref{fig:grid}, it is effortless to decode out two entities ``\textit{aching in legs}'' and ``\textit{aching in shoulders}'' by NNW relations (aching$\rightarrow$in), (in$\rightarrow$legs), and (in$\rightarrow$ shoulders), and THW relations (legs$\rightarrow$aching, Symptom) and (shoulders$\rightarrow$aching, Symptom).
Moreover, NNW and THW relations imply other effects for NER. For example, NNW relations
associate the segments of the same discontinuous entity (e.g., ``aching in'' and ``shoulders''), and they are also beneficial for identifying entity words (neighbouring) and non-entity words (non-neighbouring).
THW relations help identify the boundaries of entities, which plays an important role reported in recent NER studies \cite{zheng-etal-2019-boundary,fei2021rethinking,shen2021locate}.

\section{Unified NER Framework}

The architecture of our framework is illustrated in Figure \ref{fig:architecture}, which mainly consists of three components.
First, the widely-used pretrained language model, BERT \cite{devlin2018bert}, and bi-directional LSTM \cite{lample2016neural} are used as the encoder to yield contextualized word representations from input sentences.
Then a convolution layer is used to build and refine the representation of the word-pair grid for later word-word relation classification.
Afterward, a co-predictor layer \cite{li2021mrn} that contains a biaffine classifier and a multi-layer perceptron is leveraged for jointly reasoning the relations between all word pairs.

\subsection{Encoder Layer}

We leverage BERT \cite{devlin2018bert} as inputs for our model since it has been demonstrated to be one of the state-of-the-art models for representation learning in NER \cite{wang2021discontinuous} and relation classification \cite{li2021mrn}.
Given an input sentence $X = \{x_1, x_2, ..., x_N\}$, we convert each token or word $x_i$ into word pieces and then feed them into a pretrained BERT module.
After the BERT calculation, each sentential word may involve vectorial representations of several pieces.
Here we employ max pooling to produce word representations based on the word piece representations.
To further enhance context modeling, we follow prior work \cite{wadden2019entity,li2021span}, adopting a bi-directional LSTM \cite{lample2016neural} to generate final word representations, i.e.,
$\mathbf{H} = \{\mathbf{h}_1, \mathbf{h}_2, ..., \mathbf{h}_N\} \in \mathbb{R}^{N \times d_h}$, where $d_h$ denotes the dimension of a word representation.

\subsection{Convolution Layer}
We adopt convolution neural networks (CNNs) as the representation refiner, since CNNs are naturally suitable for 2-D convolution on the grid, and also show the very prominence on handling relation determination jobs \cite{zeng2014relation,wang2016relation}. 
Our convolution layer includes three modules, including a condition layer with normalization \cite{liu-etal-2021-modulating} for generating the representation of the word-pair grid, a BERT-style grid representation build-up to enrich the representation of the word-pair grid, and a multi-granularity dilated convolution for capturing the interactions between close and distant words.

\noindent\textbf{Conditional Layer Normalization}
Since the goal of our framework is to predict the relations between word pairs, it is important to generate a high-quality representation of the word-pair grid, which can be regarded as a 3-dimensional matrix, $\mathbf{V} \in \mathbb{R}^{N \times N \times d_h}$, where $\mathbf{V}_{ij}$ denotes the representation of the word pair $(x_i, x_j)$. Because both NNW and THW relations are directional, i.e., from a word $x_i$ in certain row to a word $x_j$ in certain column as shown in Figure \ref{fig:grid} (e.g., aching$\rightarrow$in and legs$\rightarrow$aching), the representation $\mathbf{V}_{ij}$ of the word pair $(x_i, x_j)$ can be considered as a combination of the representation $\mathbf{h}_i$ of $x_i$ and $\mathbf{h}_j$ of $x_j$, where the combination should imply that $x_j$ is conditioned on $x_i$. 
Inspired by \citet{liu-etal-2021-modulating}, we adopt the Conditional Layer Normalization (CLN) mechanism to calculate $\mathbf{V}_{ij}$:
\setlength\abovedisplayskip{2pt}
\setlength\belowdisplayskip{2pt}
\begin{align}
\mathbf{V}_{ij} &= {\rm CLN}(\mathbf{h}_i, \mathbf{h}_j) = \gamma_{ij} \odot (\frac{\mathbf{h}_j-\mu}{\sigma}) + \lambda_{ij} \,,
\label{eq:cln1}
\end{align}
where $\mathbf{h}_i$ is the condition to generate the gain parameter $\gamma_{ij}= \mathbf{W}_{\alpha} \mathbf{h}_i+\mathbf{b}_{\alpha}$ and bias $\lambda_{ij} = \mathbf{W}_{\beta} \mathbf{h}_i+\mathbf{b}_{\beta}$ of layer normalization.
$\mu$ and $\sigma$ are the mean and standard deviation across the elements of $\mathbf{h}_j$, denoted as:
\begin{align}
\mu &= \frac{1}{d_h}\sum_{k=1}^{d_h} h_{jk}, \;\;\; \sigma = \sqrt{\frac{1}{d_h} \sum_{k=1}^{d_h} (h_{jk}-\mu)^2} \,.
\label{eq:cln2}
\end{align}
where $h_{jk}$ denotes the $k$-th dimension of $\mathbf{h}_j$.

\noindent\textbf{BERT-Style Grid Representation Build-Up}
As everyone knows, the inputs of BERT \cite{devlin2018bert} consist of three parts, namely token embeddings, position embeddings and segment embeddings, which model word, position and sentential information respectively.
Motivated by BERT, we enrich the representation of the word-pair grid using a similar idea, where the tensor $\mathbf{V} \in \mathbb{R}^{N \times N \times d_h}$ represents word information, a tensor $\mathbf{E}^d \in \mathbb{R}^{N \times N \times d_{E_d}}$ represents the relative position information between each pair of words, and a tensor $\mathbf{E}^t \in \mathbb{R}^{N \times N \times d_{E_t}}$ represents the region information for distinguishing lower and upper triangle regions in the grid.
We then concatenate three kinds of embeddings
and adopt a multi-layer perceptron (MLP) to reduce their dimensions and mix these information to get the position-region-aware representation of the grid $\mathbf{C} \in \mathbb{R}^{N \times N \times d_{c}}$.
The overall process can be formulated as:
\begin{equation}
\setlength\abovedisplayskip{2pt}
\setlength\belowdisplayskip{2pt}
    \mathbf{C} = {\rm MLP_1}([\mathbf{V}; \mathbf{E}^d; \mathbf{E}^t]) \,.
\end{equation}

\noindent\textbf{Multi-Granularity Dilated Convolution}
Motivated by TextCNN \cite{kim-2014-convolutional}, we adopt multiple 2-dimensional dilated convolutions (DConv) with different dilation rates $l$ (e.g., $l \in [1, 2, 3]$) to capture the interactions between the words with different distances, because our model is to predict the relations between these words.
The calculation in one dilated convolution can be formulated as:
\begin{equation}
\setlength\abovedisplayskip{2pt}
\setlength\belowdisplayskip{2pt}
    \mathbf{Q}^l = \sigma({\rm DConv}_l(\mathbf{C})) \,,
\end{equation}
where $\mathbf{Q}^l \in \mathbb{R}^{N \times N \times d_c}$ denotes the output of the dilation convolution with the dilation rate $l$,
$\sigma$ is the GELU activation function \cite{hendrycks2016gaussian}.
After that, we can obtain the final word-pair grid representation $\mathbf{Q} = [\mathbf{Q}^1, \mathbf{Q}^2, \mathbf{Q}^3] \in \mathbb{R}^{N \times N \times 3d_c}$.

\subsection{Co-Predictor Layer}

After the convolution layer, we obtain the word-pair grid representations $\mathbf{Q}$, which are used to predict the relation between each pair of words using an MLP.
However, prior work \cite{li2021mrn} has shown that MLP predictor can be enhanced by collaborating with a biaffine predictor for relation classification.
We thus take these two predictors concurrently to calculate two separate relation distributions of word pair $(x_i, x_j)$, and combine them as the final prediction.

\noindent\textbf{Biaffine Predictor}
The input of the biaffine predictor is the output $\mathbf{H} = \{\mathbf{h}_1, \mathbf{h}_2, ..., \mathbf{h}_N\} \in \mathbb{R}^{N \times d_h}$ of the encoder layer, which can be considered as a residual connection \cite{he2016resnet} that is widely-used in current deep learning research.
Given the word representations $\mathbf{H}$, we use two MLPs to calculate the subject ($x_i$) and object ($x_j$) word  representations, $\mathbf{s}_i$ and $\mathbf{o}_j$ respectively.
Then, a biaffine classifier \cite{dozat2017deep} is used to compute the relation scores between a pair of subject and object words ($x_i$, $x_j$):
\setlength\abovedisplayskip{2pt}
\setlength\belowdisplayskip{2pt}
\begin{align}
    \mathbf{s}_i &= {\rm MLP}_2(\mathbf{h}_i) \ , \\
    \mathbf{o}_j &= {\rm MLP}_3(\mathbf{h}_j) \ , \\
    \mathbf{y}_{ij}' &= {\mathbf{s}_i}^\top \mathbf{U} \mathbf{o}_j + \mathbf{W}[\mathbf{s}_i;\mathbf{o}_j] + \mathbf{b} \ ,
\end{align}
where $\mathbf{U}$, $\mathbf{W}$ and $\mathbf{b}$ are trainable parameters,
$\mathbf{s}_i$ and $\mathbf{o}_j$ denote the subject and object representations of the $i$-th and $j$-th word, respectively. Here $\mathbf{y}_{ij}' \in \mathbb{R}^{|\mathcal{R}|}$ is the scores of the relations pre-defined in $\mathcal{R}$.

\noindent\textbf{MLP Predictor}
Based on the word-pair grid representation $\mathbf{Q}$,
we adopt an MLP to calculate relations scores for word pairs ($x_i$, $x_j$) using $\mathbf{Q}_{ij}$:
\begin{equation}
\setlength\abovedisplayskip{2pt}
\setlength\belowdisplayskip{2pt}
    \mathbf{y}_{ij}'' = {\rm MLP}(\mathbf{Q}_{ij}) \, ,
\end{equation}
where $\mathbf{y}_{ij}'' \in \mathbb{R}^{|\mathcal{R}|}$ is the scores of the relations pre-defined in $\mathcal{R}$.
The final relation probabilities $\mathbf{y}_{ij}$ for the word pair $(x_i, x_j)$ are calculated by combining the scores from the biaffine and MLP predictors:
\begin{equation}
\setlength\abovedisplayskip{2pt}
\setlength\belowdisplayskip{2pt}
    \mathbf{y}_{ij} = {\rm Softmax}(\mathbf{y}_{ij}' + \mathbf{y}_{ij}'') \ .
\end{equation}

\begin{figure}
    \centering
    \includegraphics[width=0.97\columnwidth]{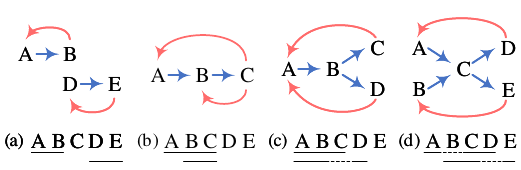}
    \caption{Four decoding cases for the word sequence ``ABCDE''. (a) ``AB'' and ``DE'' are flat entities. (b) The flat entity ``BC'' is nested in ``ABC''. (c) The entity ``ABC'' is overlapped with a discontinuous entity ``ABD''. (d) Two discontinuous entities ``ACD'' and ``BCE'' are overlapped. The blue and red arrows indicate NNW and THW relations.}
    \label{fig:decoding}
\end{figure}

\subsection{Decoding}

The predictions of our model are the words and their relations, which can be considered as a directional word graph.
The decoding object is to find certain paths from one word to anther word in the graph using NNW relations.
Each path corresponds to an entity mention.
Besides the type and boundary identification for NER, THW relations can also be used as auxiliary information for disambiguation.
Figure \ref{fig:decoding} illustrates four cases for decoding from easy to difficult.

\begin{itemize}
\setlength{\itemsep}{0pt}
\setlength{\parsep}{0pt}
\setlength{\parskip}{0pt}
  \item In the example (a), two paths ``A$\rightarrow$B'' and ``D$\rightarrow$E'' correspond to flat entities, and THW relations indicate their boundaries and types.
  \item In the example (b), if there is no THW relation, we can only find one path and thus ``BC'' is missing. 
  In contrast, with the help of THW relations, it is easy to identify that ``BC'' is nested in ``ABC'', which demonstrates the necessity of THW relations.
  \item The case (c) shows how to identify discontinuous entities. Two paths ``A$\rightarrow$B$\rightarrow$C'' and ``A$\rightarrow$B$\rightarrow$D'' can be found, and the NNW relation contributes to connecting the discontinuous spans ``AB'' and ``D''.
  \item Considering a complex and rare case (d), it is impossible to decode correct entities ``ACD'' and ``BCE'' because we can find 4 paths in this ambiguous case using only NNW relations. 
  In contrast, only using THW relations will recognize continuous entities (e.g., ``ABCD'') rather than correct discontinuous entities (e.g., ``ACD''). 
  Therefore, we can obtain correct answers by collaboratively using both relations.
\end{itemize}

\begin{table*}[ht!]
\centering
\small
\begin{tabular}{llcccccc}
\toprule
\multirow{2}{*}{} & \multirow{2}{*}{} & \multicolumn{3}{c}{CoNLL2003} & \multicolumn{3}{c}{OntoNotes 5.0} \\ \cmidrule(r){3-5} \cmidrule(r){6-8} 
& & P & R & F1 & P & R & F1 \\ \hline
\multirow{2}{*}{$\bullet$ \textbf{Sequence Labeling}} & \citet{lample2016neural} & - & - & 90.94 & - & - & - \\
 & \citet{strubell2017fast} & - & - & 90.65 & - & - & 86.84 \\ \hline
\multirow{2}{*}{$\bullet$ \textbf{Span-based}} & \citet{yu2020named} $\dag$ & \textbf{92.91} & 92.13 & 92.52 & 90.01 & 89.77 & 89.89 \\ 
 & \citet{shen2021locate} & 92.13 & \textbf{93.73} & 92.94 & - & - & - \\ \hline
$\bullet$ \textbf{Hypergraph-based} & \citet{wang2018neural} & - & - & 90.50 & - & - & - \\ \hline
\multirow{2}{*}{$\bullet$ \textbf{Seq2Seq}} & \citet{strakova2019neural} & - & - & 92.98 & - & - & - \\
 & \citet{yan2021unified} $\dag$ & 92.56 & 93.56 & 93.05 & 89.62 & 90.92 & 90.27 \\ \hline
   & W$^2$NER (ours) & 92.71 & 93.44 & \textbf{93.07} & \textbf{90.03} & \textbf{90.97} & \textbf{90.50} \\ \hline

\end{tabular}
\caption{Results for English flat NER datasets. ``$\dag$'' denotes our re-implementation via their code. We run our model for 5 times and report averaged values.\footnote[3]{}
}
\label{tab:flat-en}
\end{table*}

\newcolumntype{A}{p{0.2\textwidth}}
\newcolumntype{B}{p{0.04\textwidth}<{\centering}}
\begin{table*}[t!]
\centering
\small
\begin{tabular}{ABBBBBBBBBBBB} 
 \toprule
 \multirow{2}{*}{} & \multicolumn{3}{c}{OntoNotes 4.0} & \multicolumn{3}{c}{MSRA} & \multicolumn{3}{c}{Resume}  & \multicolumn{3}{c}{Weibo} \\ \cmidrule(r){2-4} \cmidrule(r){5-7} \cmidrule(r){8-10} \cmidrule(r){11-13} 
  & P & R & F1 & P & R & F1 & P & R & F1 & P & R & F1 \\  \hline 
\citet{zhang2018chinese} & 76.35 & 71.56 & 73.88 & 93.57 & 92.79 & 93.18
 & 94.81 & 94.11 & 94.46 & 53.04 & 62.25 & 58.79 \\
\citet{yan2019tener} & - & - & 72.43 & - & - & 92.74 & - & - & 95.00 & - & - & 58.17 \\
\citet{gui2019lexicon} & 76.40 & 72.60 & 74.45 & 94.50 & 92.93 & 93.71
 & 95.37 & 94.84 & 95.11 & 57.14 & 66.67 & 59.92 \\
\citet{li2020flat} & - & - & 81.82 & - & - & 96.09 & - & - & 95.86 & - & - & 68.55 \\
\citet{ma2019simplify} & \textbf{83.41} & 82.21 & 82.81 & 95.75 & 95.10 & 95.42 & 96.08 & 96.13 & 96.11 & \textbf{70.94} & 67.02 & 70.50 \\ \hline
W$^2$NER (ours) & 82.31 & \textbf{83.36} & \textbf{83.08} & \textbf{96.12} & \textbf{96.08} & \textbf{96.10}  & \textbf{96.96} & \textbf{96.35} &  \textbf{96.65} & 70.84 & \textbf{73.87} & \textbf{72.32} \\  \hline

\end{tabular}
\caption{Results for Chinese flat NER datasets. All the baselines are sequence labeling methods or their variations.}
\label{tab:flat-zh}
\end{table*}

\newcolumntype{C}{p{0.17\textwidth}}
\newcolumntype{D}{p{0.17\textwidth}}
\newcolumntype{E}{p{0.042\textwidth}<{\centering}}
\begin{table*}[t!]
\centering
\small
\begin{tabular}{CDEEEEEEEEE}
\toprule
\multirow{2}{*}{} & \multirow{2}{*}{} & \multicolumn{3}{c}{ACE2004} & \multicolumn{3}{c}{ACE2005} & \multicolumn{3}{c}{GENIA} \\ \cmidrule(r){3-5} \cmidrule(r){6-8} \cmidrule(r){9-11}
& & P & R & F1 & P & R & F1 & P & R & F1 \\ \hline
$\bullet$ \textbf{Sequence Labeling} & \citet{ju2018neural} & - & - & - & 74.20 & 70.30 & 72.20 & 78.50 & 71.30 & 74.70 \\ \hline
\multirow{3}{*}{$\bullet$ \textbf{Span-based}} & \citet{wang2020pyramid} & 86.08 & 86.48 & 86.28 & 83.95 & 85.39 & 84.66  & 79.45 & 78.94 & 79.19 \\
 & \citet{yu2020named} & 87.30 & 86.00 & 86.70 & 85.20 & 85.60 & 85.40 & 81.80 & 79.30 & 80.50 \\
 & \citet{shen2021locate} & \textbf{87.44} & 87.38 & 87.41 & \textbf{86.09} & 87.27 & 86.67 & 80.19 & \textbf{80.89} & 80.54 \\ \hline
$\bullet$ \textbf{Hypergraph-based} & \citet{wang2018neural} & 78.00 & 72.40 & 75.10 & 76.80 & 72.30 & 74.50 & 77.00 & 73.30 & 75.10 \\ \hline
\multirow{2}{*}{$\bullet$ \textbf{Seq2Seq}} & \citet{strakova2019neural} & - & - & 84.33 & - & - & 83.42 & - & - & 78.20 \\
 & \citet{yan2021unified} & 87.27 & 86.41 & 86.84 & 83.16 & 86.38 & 84.74  & 78.87 & 79.60 & 79.23 \\ \hline
& W$^2$NER (ours) & 87.33 & \textbf{87.71} & \textbf{87.52} & 85.03 & \textbf{88.62} & \textbf{86.79} & \textbf{83.10} & 79.76 & \textbf{81.39} \\ \hline
\end{tabular}
\caption{Results for English overlapped NER datasets.}
\label{tab:nested-en}
\end{table*}

\subsection{Learning}

For each sentence $X = \{x_1, x_2, ..., x_N\}$,
our training target is to minimize the negative log-likelihood losses with regards to the corresponding gold labels, formalized as:
\begin{equation}
\setlength\abovedisplayskip{2pt}
\setlength\belowdisplayskip{2pt}
    \mathcal{L} = - \frac{1}{N^2} \sum^N_{i=1} \sum^N_{j=1} \sum^{|\mathcal{R}|}_{r=1} \mathbf{\hat{y}}^r_{ij} \text{log} \mathbf{y}^r_{ij} ,
\end{equation}
where $N$ it the number of words in the sentence, $\mathbf{\hat{y}}_{ij}$ is the binary vector that denotes the gold relation labels for the word pair $(x_i, x_j)$, and $\mathbf{y}_{ij}$ are the predicted probability vector.
$r$ indicates the $r$-th relation of the pre-defined relation set $\mathcal{R}$.

\section{Experimental Settings}

\footnotetext[3]{The results in Table 2-6 are also the averaged values.}

\subsection{Datasets}

To evaluate our framework for three NER subtasks, we conducted experiments on 14 datasets.

\noindent\textbf{Flat NER Datasets}
We adopt CoNLL-2003 \cite{sang2003introduction} and OntoNotes 5.0 \cite{pradhan2013towards} in English,
OntoNotes 4.0 \cite{weischedel2011ontonotes}, 
MSRA \cite{levow2006third}, Weibo \cite{peng2015named,he2016f}, and Resume \cite{zhang2018chinese} in Chinese. 
We employ the same experimental settings in previous work \cite{lample2016neural,yan2021unified,ma2019simplify,li2020flat}.

\noindent\textbf{Overlapped NER Datasets}
We conduct experiments on ACE 2004 \cite{doddington2004automatic}, 
ACE 2005 \cite{walker2011ace}, 
GENIA \cite{kim2003genia}.
For GENIA, we follow \citet{yan2021unified} to use five types of entities and split the train/dev/test as 8.1:0.9:1.0.
For ACE 2004 and ACE 2005 in English, we use the same data split as \citet{lu2015joint,yu2020named}.
For ACE 2004 and ACE 2005 in Chinese, we split the train/dev/test as 8.0:1.0:1.0.

\noindent\textbf{Discontinuous NER Datasets}
We experiment on three datasets for discontinuous NER, namely CADEC \cite{karimi2015cadec}, ShARe13 \cite{pradhan2013task} and ShARe14 \cite{mowery2014task}, all of which are derived from biomedical or clinical domain documents.
We use the preprocessing scripts provided by \citet{dai2020effective} for data splitting.
Around 10\% of entities in these datasets are discontinuous.

\begin{table*}[ht!]
\centering
\small
\begin{tabular}{CDEEEEEEEEE}
\toprule
\multirow{2}{*}{} & \multirow{2}{*}{} & \multicolumn{3}{c}{CADEC} & \multicolumn{3}{c}{ShARe13} & \multicolumn{3}{c}{ShARe14} \\ \cmidrule(r){3-5} \cmidrule(r){6-8} \cmidrule(r){9-11}
& & P & R & F1 & P & R & F1 & P & R & F1 \\ \hline
$\bullet$ \textbf{Sequence Labeling} & \citet{tang2018recognizing} & 67.80 & 64.99 & 66.36 & - & - & - & - & - & - \\ \hline
$\bullet$ \textbf{Span-based} & \citet{li2021span} & - & - & 69.90 & - & - & 82.50 & - & - & - \\  \hline
$\bullet$ \textbf{Hypergraph-based} & \citet{wang2019combining} & 72.10 & 48.40 & 58.00 & 83.80 & 60.40 & 70.30 & 79.10 & 70.70 & 74.70 \\ \hline
\multirow{2}{*}{$\bullet$ \textbf{Seq2Seq}} & \citet{yan2021unified} & 70.08 & 71.21 & 70.64 & 82.09 & 77.42 & 79.69 & 77.20 & 83.75 & 80.34 \\ 
& \citet{fei2021rethinking} & \textbf{75.50} & 71.80 & 72.40 & \textbf{87.90} & 77.20 & 80.30 & - & - & - \\ \hline
\multirow{2}{*}{$\bullet$ \textbf{Others}} & \citet{dai2020effective} & 68.90 & 69.00 & 69.00 & 80.50 & 75.00 & 77.70 & 78.10 & 81.20 & 79.60 \\
 & \citet{wang2021discontinuous} & 70.50 & \textbf{72.50} & 71.50 & 84.30 & 78.20 & 81.20 & 78.20 & \textbf{84.70} & 81.30 \\ \hline
& W$^2$NER (ours) & 74.09 & 72.35 & \textbf{73.21} & 85.57 & \textbf{79.68} & \textbf{82.52} & \textbf{79.88} & 83.71 & \textbf{81.75} \\ \hline
\end{tabular}
\caption{Results for discontinuous NER datasets.\footnote[4]{}
}
\label{tab:discontinuous-en}
\end{table*}

\begin{table}[t!]
\small
\centering
\begin{tabular}{lcc}
\toprule
 & ACE2004 & ACE2005 \\ \hline
\citet{yu2020named} $\star$ & 87.35 & 88.39 \\
\citet{shen2021locate} $\star$ & 87.47 & 88.21 \\ \hline
W$^2$NER (ours) & \textbf{88.00} & \textbf{88.81} \\ \hline
\end{tabular}
\caption{F1s for Chinese overlapped NER datasets. 
Models with ``$\star$'' are adapted to target datasets using their code.}
\label{tab:nested-zh}
\end{table}

\subsection{Baselines}
\textbf{Tagging-based methods}, which assign a tag to every token with different label schemes, such as BIO \cite{lample2016neural}, BIOHD \cite{tang2018recognizing}, and BIEOS \cite{li2020flat,ma2019simplify}.
\textbf{Span-based methods}, which enumerate all possible spans and combine them into entities \cite{yu2020named,li2021span}.
\textbf{Hypergraph-based approaches}, which utilize hypergraphs to represent and infer entity mentions \cite{lu2015joint,wang2018neural,katiyar2018nested}.
\textbf{Seq2Seq methods}, which generate entity label sequences \cite{strubell2017fast}, index or word sequences \cite{yan2021unified,fei2021rethinking} at the decoder side.
\textbf{Other methods}, which is different from the methods above, such as transition-based \cite{dai2020effective} and clique-based \cite{wang2021discontinuous} approaches.

\section{Experimental Results}

\subsection{Results for Flat NER}

We evaluate our framework on six datasets.
As shown in Table \ref{tab:flat-en},
Our model achieves the best performances with 93.07\% F1 and 90.50\% F1 on CoNLL 2003 and OntoNotes 5.0 datasets.
Especially, our model outperforms another unified NER framework \citet{yan2021unified} by 0.23\% in terms of F1 on OntoNotes 5.0. 
The results in Chinese datasets are shown in Table \ref{tab:flat-zh}, where baselines are all tagging-based methods.
We find that our model outperforms the previous SoTA results by 0.27\%, 0.01\%, 0.54\% and 1.82\% on OntoNotes 4.0, MSRA, Resume and Weibo.

\begin{figure}[t]
\centering
\begin{tikzpicture}
  \begin{axis}[name=plot1,ybar=7pt,height=0.45\columnwidth,width=0.57\columnwidth,ylabel=F1 (\%),xlabel=(a) Overlapped NER,xtick style={draw=none},enlargelimits=0.5,xticklabel style = {yshift=5pt},ylabel shift=-3pt,xticklabels={,,},compat=newest,
  legend style={at={(1.15,1.53)},draw=none, font=\small,
        anchor=north,legend columns=2,
        /tikz/every even column/.append style   = {column sep=0.5cm, text width=7em},
                 /tikz/every odd column/.append style    = {column sep=0.15cm,}
                 }
                 ,xlabel style={at={(0.45,-0.05)}}]
    \addplot[blue,fill=blue!30!white] coordinates {(0,56.91)};
    \addplot[black,fill=gray] coordinates {(0,58.94)};
    \addplot[brown!60!black,fill=brown!30!white] coordinates {(0,61.54)};
    \addplot[red,fill=red!30!white] coordinates {(0,66.07)};
    \legend{\citet{dai2020effective},\citet{yan2021unified}, \citet{wang2021discontinuous},Ours}
  \end{axis}
  \begin{axis}[name=plot2,at={($(plot1.east)+(0.7cm,0)$)},anchor=west,ybar=7pt,height=0.45\columnwidth,width=0.57\columnwidth,xtick style={draw=none},enlargelimits=0.5,xticklabel style = {yshift=5pt},ylabel shift=-5pt,xticklabels={,,},compat=newest,xlabel=(b) Discontinuous NER,xlabel style={at={(0.45,-0.05)}}]
    \addplot[blue,fill=blue!30!white] coordinates {(0,59.47)};
    \addplot[black,fill=gray] coordinates {(0,59.68)};
    \addplot[brown!60!black,fill=brown!30!white] coordinates {(0,61.40)};
    \addplot[red,fill=red!30!white] coordinates {(0,65.20)};
  \end{axis}
\end{tikzpicture}
\caption{Results of overlapped (a) and discontinuous mentions (b) on ShARe14.
}
\label{fig:effect}
\end{figure}
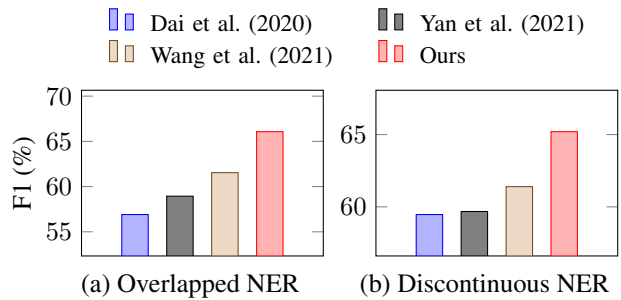

\footnotetext[4]{Note that discontinuous NER datasets include both flat and overlapped entities as well.}

\subsection{Results for Overlapped NER}

Table \ref{tab:nested-en} presents the results for three overlapped NER datasets in English.
Our W$^2$NER model outperforms the previous works, including tagging-based \cite{ju2018neural}, span-based \cite{wang2020pyramid,yu2020named,shen2021locate}, hypergraph-based \cite{wang2018neural} and sequence-to-sequence \cite{strakova2019neural,yan2021unified} approaches,
and achieves the SoTA performances on F1 scores, with 87.52\%, 86.79\% and 81.39\% on ACE2004, ACE2005 and GENIA, respectively.
For ACE2004 and ACE2005 corpora in Chinese, we reproduce the SoTA models proposed by \citet{yu2020named} and \citet{shen2021locate},
and list their results in Table \ref{tab:nested-zh}.
Our model can significantly outperform the two baselines by 0.53\% and 0.42\%.

\subsection{Results for Discontinuous NER}

Table \ref{tab:discontinuous-en} presents the comparisons between our model and other baselines in three discontinuous NER datasets.
As seen, our model outperforms previous best model \cite{fei2021rethinking,wang2021discontinuous} by 0.81\%, 0.02\%, and 0.45\% in F1s in the CADEC, ShARe13 and ShARe14 datasets, respectively,
leading to new SoTA results.

Since the above datasets also include flat entities, we further investigate the performances of our model on recognizing only overlapped or discontinuous entities,
as shown in Figure \ref{fig:effect}.
We can learn that the clique-based model \cite{wang2021discontinuous} shows better performances than the Seq2Seq model \cite{yan2021unified} and transition-based method \cite{dai2020effective}.
Most importantly, our system achieves the best results against all other baselines for both overlapped and discontinuous NER.

\begin{table}[t!]
\small
\centering
\begin{tabular}{llll}
\toprule
                  & CoNLL2003 & ACE2005 & CADEC \\ \hline
Ours              & \textbf{93.07}   & \textbf{86.79}   & \textbf{73.21}  \\ \hline
 - Region Emb.  & 92.80\ \underline{\scriptsize (-0.27)}   & 86.39\ \underline{\scriptsize(-0.40)}   & 72.56\ \underline{\scriptsize (-0.65)}  \\
 - Distance Emb. & 92.89\ {\scriptsize(-0.18)}   & 86.47\ {\scriptsize(-0.32)}   & 72.66\ {\scriptsize(-0.55)}  \\ \hdashline
 - All DConv        & 92.31\ {\scriptsize(-0.76)}   & 86.07\ {\scriptsize(-0.72)}   & 72.45\ {\scriptsize(-0.76)} \\
 - DConv($l$=1) & 93.05\ {\scriptsize(-0.02)}   & 86.64\ {\scriptsize(-0.15)}   & 73.12\ {\scriptsize(-0.09)}  \\
 - DConv($l$=2) & 92.78\ \underline{\scriptsize(-0.29)}   & 86.58\ \underline{\scriptsize(-0.21)}   & 72.95\ \underline{\scriptsize(-0.26)}  \\
 - DConv($l$=3) & 92.82\ {\scriptsize(-0.25)}   & 86.59\ {\scriptsize(-0.20)}   & 73.10\ {\scriptsize(-0.11)}  \\ \hdashline
 - Biaffine    & 93.02\ {\scriptsize(-0.05)}   & 86.30\ {\scriptsize(-0.49)}   & 72.71\ {\scriptsize(-0.50)}  \\
 - MLP        & 91.87\ \underline{\scriptsize(-1.20)}   & 85.66\ \underline{\scriptsize(-1.13)}   & 68.04\ \underline{\scriptsize(-5.17)} \\ \hdashline
 - NNW        & 92.65\ {\scriptsize(-0.42)}   & 86.23\ {\scriptsize(-0.56)}   & 69.01\ \underline{\scriptsize(-4.20)} \\ \hline

\end{tabular}
\caption{Model ablation studies (F1s). DConv($l$=1) denots the convolution with the dilation rate 1.}
\label{tab:ablation}
\end{table}

\subsection{Model Ablation Studies}

We ablate each part of our model on the CoNLL2003, ACE2005 and CADEC datasets, as shown in Table \ref{tab:ablation}.
First, without region and distance embeddings, we observe slight performance drops on the three datasets.
By removing all convolutions, the performance also drops obviously,
which verifies the usefulness of the multi-granularity dilated convolution.
Furthermore, after removing convolutions with different dilation rate, the performance also decreases, especially for the convolution with the dilation rate 2.

Comparing the biaffine and MLP in the co-predictor layer, we find that although the MLP plays a leading role, the biaffine also brings about 0.5\% gains at most. 
At last, when the NNW relation is removed, the F1s on all datasets drop, especially on the CADEC (4.2\%).
This is because the CADEC dataset also contains discontinuous entities and without the NNW relation, discontinuous spans will be incorrectly recognized as continuous ones, as shown in Figure \ref{fig:decoding}(d).
Therefore, the results of ablation studies on the NNW relation demonstrate its importance as we argued before.

\section{Related Work on NER}

\noindent\textbf{Sequence Labeling Approaches}
NER is usually considered as a sequence labeling problem, to assign each token a tag from a pre-designed tagging scheme (e.g., \emph{BIO}).
Current mainstream work combine the CRF \cite{lafferty2001conditional,finkel2005incorporating} with neural architecture, such as CNN \cite{collobert2011natural,strubell2017fast}, bi-directional LSTM \cite{huang2015bidirectional,lample2016neural}, and Transformer \cite{yan2019tener,li2020flat}.
However, these methods fail to directly solve neither overlapped nor discontinuous NER.
\citet{ju2018neural} propose a neural model for nested NER by dynamically stacking flat NER layers.
\citet{tang2018recognizing} extend the BIO label scheme to BIOHD to address the problem of discontinuous mention.

\noindent\textbf{Span-based Approaches}
There have been several studies that cast NER as span-level classification, i.e., enumerating all possible spans, and determining if they are valid mentions and the types \cite{xu2017local,luan2019general,yamada2020luke}.
\citet{yu2020named} utilize biaffine attention \cite{dozat2017deep} to measure the possibility as a mention of a text span.
\citet{li2020unified} reformulate NER as a machine reading comprehension (MRC) task and extract entities as the answer spans.
\citet{shen2021locate} implement a two-stage identifier to generate span proposals through a filter and a regressor, and then classify them into the corresponding categories.
\citet{li2021span} convert the discontinuous NER to find complete subgraphs from a span-based entity fragment graph, and achieve competitive results.
But, due to the exhaustively enumerating nature, those methods suffer from maximal span lengths and considerable model complexity, especially for long-span entities.

\noindent\textbf{Hypergraph-based Approaches}
\citet{lu2015joint} first propose the hypergraph model for overlapped NER, by exponentially representing possible mentions.
The method is then widely explored by follow-up work \cite{muis2016learning,katiyar2018nested,wang2018neural}.
For instance, \citet{muis2016learning} extend the method for discontinuous NER,
and \citet{wang2018neural} utilize deep neural networks to enhance the hypergraph model.

\noindent\textbf{Sequence-to-Sequence Approaches}
\citet{gillick2015multilingual} first apply the Seq2Seq model for NER, taking as inputs the sentence, and outputting all the entity start positions, span lengths and labels.
\citet{strakova2019neural} use the Seq2Seq architecture for overlapped NER with enhanced BILOU scheme.
\citet{fei2021rethinking} employ Seq2Seq with pointer network for discontinuous NER.
The latest attempt in \cite{yan2021unified} tackles the unified NER via a Seq2Seq model with pointer network based-on BART \cite{lewis2020bart}, generating a sequence of all possible entity start-end indexes and types.
Seq2Seq architecture unfortunately suffers from the potential decoding efficiency problem as well as the exposure bias issue.

\noindent\textbf{Differences between Our Approach and Previous Approaches}
Most of the existing NER work mainly consider more accurate entity boundary identification.
In this work, we explore a different task modeling for unified NER, i.e., a formalism as word-word relation classification.
Our method can effectively model the relations between both the boundary-words and inside-words of entities.
Also, our method with 2D grid-tagging can substantially avoid the drawbacks in current best-performing baselines, e.g., span-based and sequence-to-sequence models.

\section{Conclusion}

In this paper, we propose a novel unified NER framework based on word-word relation classification to address unified NER concurrently.
The relations between word pairs are pre-defined as next-neighboring-word relations and tail-head-word relations.
We find that our framework is quite effective for various NER, which achieves SoTA performances for 14 widely-used benchmark datasets.
Moreover, we propose a novel backbone model that consists of a BERT-BiLSTM encoder layer, a convolution layer for building and refining the representation of the word-pair grid, and a co-predictor layer for jointly reasoning relations. 
Through ablation studies, we find that our convolution-centric model performs well and several proposed modules such as the co-predictor and grid representation enrichment are also effective.
Our framework and model are easy to follow, which will promote the development of NER research.

\newpage

\section*{Acknowledgments}

This work is supported by the National Natural Science Foundation of China (No.61772378, No. 62176187), 
the National Key Research and Development Program of China (No. 2017YFC1200500), 
the Research Foundation of Ministry of Education of China (No.18JZD015),
the Youth Fund for Humanities and Social Science Research of Ministry of Education of China (No. 22YJCZH064).
This work is also the research result of the independent scientific research project (humanities and social sciences) of Wuhan University, supported by the Fundamental Research Funds for the Central Universities.

\bibliography{aaai22}

\clearpage

\appendix

\section{Decoding}

The decoding procedure is summarized in Algorithm \ref{alg:algorithm}.
The relationships $R$ of all the word pairs serve as the inputs.
The decoding object is to find all the entity word index sequences with their corresponding categories.
We first select all the THW-\texttt{*} relations in the lower triangle region of the word-pair grid (lines 2-3).
For the entities containing only one token, 
we can decode them out just using THW relations (lines 5-7).
For other entities, we construct a graph, in which nodes are words and edges are NNW relations.
Then we use the deep first search algorithm to find all the paths from the head word to the tail word, which are the word index sequences of corresponding entities (lines 9-10).
In Algorithm \ref{alg:algorithm}, we define a function ``Track'' to perform such deep first path search (lines 12-21).

\begin{algorithm}[t]
\caption{Decoding algorithm.}
\label{alg:algorithm}
\textbf{Input}: The relations $R$ of all the word pairs. $R_{ij}$ is the relation of word pair $(x_i, x_j)$, where $i, j \in [1, N]$. \\
\textbf{Output}: The list of entities with their word index sequence set $E$ and label set $T$.
\begin{algorithmic}[1] 
\STATE $E=[]$, $T = []$.
\FOR{$R_{ij} \in R$ and $i \ge j$}
\IF{$isTHWrelation(R_{ij})$}
\STATE $S = [j]$.
\IF{$i = j$}
\STATE $E.add(S)$
\STATE $T.add(t)$
\ELSE
\FOR{$k \in (j, N]$}
\STATE $Track(S, R_{jk}, k, i, R_{ij})$
\ENDFOR
\ENDIF
\ENDIF
\ENDFOR
\STATE \textbf{return} $E$, $T$
\STATE \textbf{function} $Track(S, r, m, e, t)$
\STATE \quad \textbf{if} $isNNWrelation(r)$ and $m \leq e$ \textbf{then}
\STATE \quad \quad $S.add(m)$
\STATE \quad \quad \textbf{if} $m = e$ \textbf{then}
\STATE \quad \quad \quad $E.add(S)$
\STATE \quad \quad \quad $T.add(t)$
\STATE \quad \quad \textbf{else}
\STATE \quad \quad \quad \textbf{for} $k \in (m, N]$ \textbf{do}
\STATE \quad \quad \quad \quad $Track(S, R_{mk}, k, e, t)$
\STATE \quad \quad $S.pop()$
\end{algorithmic}
\end{algorithm}

\section{Implementation Details}

\begin{table}[t]
    \centering
    \begin{tabular}{lc}
    \hline
        Hyper-parameter & Value \\ \hline
        $d_h$ & [768, 1024] \\
        $d_{E_d}$ & 20 \\
        $d_{E_t}$ & 20 \\
        $d_c$ & [64, 96, 128] \\
        Dropout & 0.5 \\
        Learning rate (BERT) & [1e-5, 5e-6] \\
        Learning rate (others) & 1e-3 \\
        Batch size & [8, 12, 16] \\ \hline
    \end{tabular}
    \caption{Hyper-parameter settings.}
    \label{tab:settings}
\end{table}

In this section, we provide more details of our experiments.
Hyper-parameter settings are listed in Table \ref{tab:settings}.
Considering the domains of the datasets, we employ BioBERT \cite{lee2020biobert} for GENIA and CADEC, Clinical BERT \cite{alsentzer2019publicly} for ShARe 13 and 14, and vanilla BERT \cite{devlin2018bert} for the other datasets.
We adopt AdamW \cite{loshchilov2019decoupled} optimizer.
Our model is implemented with PyTorch and trained with a NVIDIA RTX 3090 GPU.
All the hyper-parameters are tuned on the development set.

\section{Evaluation Metrics}

In terms of evaluation metrics, we follow prior work \cite{lu2015joint,yu2020named, yan2021unified} and employ the precision (P), recall (R) and F1-score (F1).
A predicted entity is counted as true-positive if its token sequence and type match those of a gold entity.
We run each experiment for 5 times and report the averaged value.

\section{Efficiency Comparisons}

\begin{table}[t]
\centering
\begin{tabular}{lccc}
\hline
\multirow{2}{*}{Model} & \multirow{2}{*}{\#Param.} & Training & Inference \\
                       &                           & (sent/s) & (sent/s)  \\ \hline
\citet{dai2020effective}   & 102.2M & 24.7     & 66.5      \\
\citet{yan2021unified} & 408.4M & 63.6     & 19.2      \\
\citet{wang2021discontinuous}   & 116.7M     & 39.3     & 109.7     \\
W$^2$NER (ours)    & 112.3M & 116.1    & 365.7    \\ \hline
\end{tabular}

\caption{Parameter number and running speed comparisons on CADEC.}
\label{tab:speed}
\end{table}

Table \ref{tab:speed} lists the parameter numbers and running speeds during training and inference of three baselines and our model.
For fair comparison, all of these models are implemented using PyTorch and tested using the NVIDIA RTX 3090 GPU.
First, we can see that the Seq2Seq model \cite{yan2021unified} has around 4 times of parameters more than the other three models, due to the utilization of the Seq2Seq pre-training model, BART-Large \cite{lewis2020bart}.
Furthermore, the training and inference speeds of our model are about 5 times faster than the transition-based model \cite{dai2020effective} and 3 times faster than the span-based model \cite{wang2021discontinuous}, which verify the efficiency of our model. In other words, our model leverages less parameters but achieves better performances and faster training and inference speeds.

\section{Supplemental Experiments for Recognizing Overlapped or Discontinuous Entities}

\begin{figure}[t]
\centering
\begin{tikzpicture}
  \begin{axis}[name=plot1,ybar=7pt,height=0.45\columnwidth,width=0.57\columnwidth,ylabel=F1 (\%) on CADEC,xlabel=(a) Overlapped NER,xtick style={draw=none},enlargelimits=0.5,xticklabel style = {yshift=5pt},ylabel shift=-3pt,xticklabels={,,},compat=newest,
  legend style={at={(1.15,1.46)},draw=none, font=\small,
        anchor=north,legend columns=2,
        /tikz/every even column/.append style   = {column sep=0.5cm, text width=7em},
                 /tikz/every odd column/.append style    = {column sep=0.15cm,}
                 }
                 ,xlabel style={at={(0.45,-0.05)}}]
    \addplot[blue,fill=blue!30!white] coordinates {(0,64.05)};
    \addplot[black,fill=gray] coordinates {(0,65.32)};
    \addplot[brown!60!black,fill=brown!30!white] coordinates {(0,70.40)};
    \addplot[red,fill=red!30!white] coordinates {(0,72.86)};
    \legend{\citet{dai2020effective},\citet{yan2021unified}, \citet{wang2021discontinuous},Ours}
  \end{axis}
  \begin{axis}[name=plot2,at={($(plot1.east)+(0.7cm,0)$)},anchor=west,ybar=7pt,height=0.45\columnwidth,width=0.57\columnwidth,xtick style={draw=none},enlargelimits=0.5,xticklabel style = {yshift=5pt},ylabel shift=-5pt,xticklabels={,,},compat=newest,xlabel=(b) Discontinuous NER,xlabel style={at={(0.45,-0.05)}}]
    \addplot[blue,fill=blue!30!white] coordinates {(0,64.85)};
    \addplot[black,fill=gray] coordinates {(0,64.18)};
    \addplot[brown!60!black,fill=brown!30!white] coordinates {(0,68.22)};
    \addplot[red,fill=red!30!white] coordinates {(0,70.29)};
  \end{axis}
    \begin{axis}[name=plot3,at={($(plot1.south)-(0,1cm)$)},anchor=north,ybar=7pt,height=0.45\columnwidth,ylabel=F1 (\%) on ShARe13,width=0.57\columnwidth,xtick style={draw=none},enlargelimits=0.5,xticklabel style = {yshift=5pt},ylabel shift=-5pt,xticklabels={,,},compat=newest,xlabel=(c) Overlapped NER,xlabel style={at={(0.45,-0.05)}}]
    \addplot[blue,fill=blue!30!white] coordinates {(0,53.86)};
    \addplot[black,fill=gray] coordinates {(0,58.81)};
    \addplot[brown!60!black,fill=brown!30!white] coordinates {(0,61.60)};
    \addplot[red,fill=red!30!white] coordinates {(0,65.85)};
  \end{axis}
    \begin{axis}[name=plot4,at={($(plot2.south)-(0,1cm)$)},anchor=north,ybar=7pt,height=0.45\columnwidth,width=0.57\columnwidth,xtick style={draw=none},enlargelimits=0.5,xticklabel style = {yshift=5pt},ylabel shift=-5pt,xticklabels={,,},compat=newest,xlabel=(d) Discontinuous NER,xlabel style={at={(0.45,-0.05)}}]
    \addplot[blue,fill=blue!30!white] coordinates {(0,55.40)};
    \addplot[black,fill=gray] coordinates {(0,57.91)};
    \addplot[brown!60!black,fill=brown!30!white] coordinates {(0,61.44)};
    \addplot[red,fill=red!30!white] coordinates {(0,66.13)};
  \end{axis}
\end{tikzpicture}

\caption{Results of overlapped (a) and discontinuous mentions (b) on CADEC, and overlapped (c) and discontinuous mentions (d) on ShARe13.}
\label{fig:effect_appendix}

\end{figure}

\begin{table*}[!t]
\centering
\small
\begin{tabular}{clccccccccc}
\hline
\multicolumn{2}{l}{\multirow{2}{*}{}}          & \multicolumn{5}{c}{Sentence}                 & \multicolumn{4}{c}{Mention}          \\ \cmidrule(r){3-7} \cmidrule(r){8-11}
\multicolumn{2}{l}{}                           & \#All & \#Train & \#Dev & \#Test & \#Avg.Len & \#All & \#Ovlp. & \#Dis. & \#Avg.Len \\ \hline
\multirow{6}{*}{Flat NER}       & CoNLL2003     & 20,744 & 17,291 & - & 3,453 & 14.38 & 35,089 & - & - & 1.45  \\
                               & OntoNotes 5.0 & 76,714 & 59,924 & 8,528 & 8,262 & 18.11 & 104,151 & - & - & 1.83  \\ \cdashline{2-11}
                               & OntoNotes 4.0 & 24,393 & 15,736 & 4,306 & 4,351 & 36.92 & 28,006 & - & - & 3.02 \\
                               & MSRA          & 50,847 & 46,471 & - & 4,376 & 45.54 & 80,884 & - & - & 3.24 \\
                               & Resume        & 4,759 & 3,819 & 463 & 477 & 32.17 & 16,565 & - & - & 5.88 \\
                               & Weibo         & 1,890 & 1,350 & 270 & 270 & 54.57 & 2,689 & - & - & 2.60 \\ \hline
\multirow{5}{*}{Overlapped NER}    & ACE2004-EN       & 8,512 & 6,802 & 813 & 897 & 20.12 & 27,604 & 12,626 & - & 2.50 \\
                               & ACE2005-EN       & 9,697 & 7,606 & 1,002 & 1,089 & 17.77 & 30,711 & 12,404 & - & 2.28 \\
                               & GENIA         & 18,546 & 15,023 & 1,669 & 1,854 & 25.41 & 56,015 & 10,263 & - & 1.97 \\ \cdashline{2-11}
                               & ACE2004-ZH    & 7,325 & 5,754 & 721 & 850 & 44.92 & 33,162 & 15,219 & - & 4.05 \\
                               & ACE2005-ZH    & 7,276 & 5,876 & 660 & 740 & 42.49 & 34,150 & 15,734 & - & 4.31 \\ \hline
\multirow{3}{*}{Discontinuous NER} & CADEC         & 7,597 & 5,340 & 1,097 & 1,160 & 16.18 & 6,316 & 920 & 679 & 2.72 \\
                               & ShARe13       & 18,767 & 8,508 & 1,250 & 9,009 & 14.86 & 11,148 & 663 & 1,088 & 1.82 \\
                               & ShARe14       & 34,614 & 17,404 & 1,360 & 15,850 & 15.06 & 19,070 & 1,058 & 1,656 & 1.74 \\ \hline
\end{tabular}
\caption{Dataset Statistics. ``\#'' denotes the amount. ``Ovlp.'' and ``Dis.'' denote overlapped and discontinuous mentions, respectively.}
\label{tab:statistics}
\end{table*}

In the experiments of the manuscript, we have already shown that our model achieves better results on recognizing overlapped and discontinuous entities in the ShARe14 dataset. 
Due to page limitation, we show the performances of our model on the CADEC and ShARe13 datasets in Figure \ref{fig:effect_appendix}.
As seen, our model still ranks the first in the two datasets, demonstrating the superiority of our model on recognizing overlapped or discontinuous entities.
These experiments further demonstrate that our motivation for modeling both boundary-word relations and inside-word relations is successful.

\section{Dataset Statistics}

We evaluate our framework for three NER subtasks on 8 English datasets and 6 Chinese datasets.
In Table \ref{tab:statistics}, we present the detailed statistics of 14 datasets, including CoNLL-2003 and OntoNotes 5.0 for English flat NER, OntoNotes 4.0, MSRA, Weibo, and Resume for Chinese flat NER, ACE 2004, ACE2005, and GENIA for English overlapped NER, ACE 2004 and ACE 2005 for Chinese overlapped NER, CADEC, ShARe13, and ShARe14 for English discontinuous NER.
Especially, the three discontinuous NER datasets include all three kinds of entities.

\clearpage






\end{CJK}
\end{document}